# Automatic Information Extraction From Employment Tribunal Judgements Using Large Language Models

*Joana Ribeiro de Faria*[*], *Huiyuan Xie*[†] *and Felix Steffek*[‡]

**Abstract**

Court transcripts and judgments are rich repositories of legal knowledge, detailing the intricacies of cases and the rationale behind judicial decisions. The extraction of key information from these documents provides a concise overview of a case, crucial for both legal experts and the public. With the advent of large language models (LLMs), automatic information extraction has become increasingly feasible and efficient. This paper presents a comprehensive study on the application of GPT-4, a large language model, for automatic information extraction from UK Employment Tribunal (UKET) cases. We meticulously evaluated GPT-4's performance in extracting critical information with a manual verification process to ensure the accuracy and relevance of the extracted data. Our research is structured around two primary extraction tasks: the first involves a general extraction of eight key aspects that hold significance for both legal specialists and the general public, including the facts of the case, the claims made, references to legal statutes, references to precedents, general case outcomes and corresponding labels, detailed order and remedies and reasons for the decision. The second task is more focused, aimed at analysing three of those extracted features, namely facts, claims and outcomes, in order to facilitate the development of a tool capable of predicting the outcome of employment law disputes. Through our analysis, we demonstrate that LLMs like GPT-4 can obtain high accuracy in legal information extraction, highlighting the potential of LLMs in revolutionising the way legal information is processed and utilised, offering significant implications for legal research and practice.

[*] University of Cambridge, Faculty of Law; Newnham College.
[†] University of Cambridge, Faculty of Law and Department of Computer Science and Technology; Downing College.
[‡] University of Cambridge, Faculty of Law; Newnham College; University of Notre Dame, Law School.

This project was supported with funding from the Cambridge Centre for Data-Driven Discovery and Accelerate Programme for Scientific Discovery, made possible by a donation from Schmidt Futures.

# 1 Introduction

The advent of large language models (LLMs) has marked a significant milestone in the evolution of artificial intelligence, extending its influence across various domains, including the legal sector. These sophisticated models have redefined the possibilities within the realm of legal artificial intelligence, offering innovative solutions to complex tasks that were traditionally the preserve of human experts. The application of LLMs in legal contexts promises to streamline operations, enhance accuracy and democratise access to legal knowledge.

LLMs have been instrumental in performing a range of tasks within the legal domain, such as contract analysis and case prediction (Jayakumar et al. 2023; Parizi et al. 2023; Bernsohn et al. 2024). One of the most promising applications of LLMs in law is automatic information extraction from legal documents. This capability not only simplifies the process of analysing vast quantities of legal texts but also paves the way for sophisticated downstream applications, including predictive analytics and case summarisation.

The potential advantages of deploying LLMs for information extraction in legal contexts are manifold. Primarily, these models offer cost efficiency by reducing the need for extensive human labour in sifting through and interpreting complex legal documents. LLMs can process and extract pertinent information from a large volume of cases at a pace unattainable by human counterparts. However, the reliance on LLMs is not without its challenges. Concerns regarding the accuracy of extracted information, potential biases and the models' ability to navigate the nuanced and context-dependent nature of legal language pose significant hurdles.

Given the burgeoning interest in the intersection of artificial intelligence and law, we have chosen the United Kingdom Employment Tribunal (UKET) as our focal area of study. The UKET presents a fertile ground for exploration due to its extensive repository of publicly available judgments, which encapsulate a wide array of legal issues and decisions. By harnessing the capabilities of GPT-4, one of the most advanced LLMs, we embarked on a project to automatically extract critical information from a sample of UKET judgments. Through precise prompt engineering and manual quality checks, we aimed to ensure the reliability and relevance of our study on the extracted data.

Our research is structured around two principal extraction tasks. The first task involves a broad extraction of eight key aspects that are of general importance to both legal experts and the lay public, including: the *facts* of the case, the *claims* made by the parties, *references to legal statutes*, *references to precedents*, *general case outcomes*, *general case outcomes summarised as one of four labels* (namely "claimant wins", "claimant partially wins", "claimant loses" and "other"), *detailed order and remedies* and, finally, *reasons* for the decision. The second task is more focused, aimed at analysing three of those extracted features, namely the *facts*, *claims* and *general case outcomes (labelled)*, in order to facilitate the development of a tool capable of predicting the outcome of employment law disputes.

The paper is structured as follows. First, we introduce the UKET system, highlighting the simplicity of the judicial procedure and the clear structure of its judgments (section 2.1.). The following subsection demonstrates the extent to which UKET judgments may be determined by substantive and/or procedural factors, and introduces the concepts of procedural and substantive predictability, as opposing perspectives which highly impact the dataset for prediction models (section 2.2). Second, we provide a brief insight into how information extraction and predictive systems have been employed in the legal industry, and into the initial limitations of these systems, which have been increasingly surpassed as machine learning progresses. As many systems are provided by private companies, a public debate regarding how LLM's operate in the legal field is limited. Our paper aims to address this gap by openly introducing the data and the design of an information extraction system for UKET judgments, accompanied by a discussion of the prompt engineering developed which already sheds some light onto limitations of legal extraction systems (section 4). That is, we identify some pitfalls of the extraction system which were solved through better prompt engineering and disclose some limitations which could not be addressed despite substantial prompt enhancing attempts. Thereafter, we present the quality assessment which was used to verify the accuracy of the outputs (the first part of the quality check), and to verify the relevance of the judgments for prediction (the second part of the quality check) (section 5.1). The second part of the quality check introduces the reader to a further discussion regarding the process of development of predictive systems. It highlights that GPT-4, despite being able to accurately extract information from case files, may not be able – in the absence of a carefully drafted predictive prompt – to identify the judgments which allow for substantive predictability. Section 5.2. quantitatively assesses the results produced by GPT-4 in our study and is followed by a qualitative discussion which highlights the potential of GPT-4 and the limitations that it still has for extraction of information (section 5.3). Finally, section 5.4, taking into consideration how GPT-4 extracts and presents information, discusses whether it is feasible to develop a predictive system based on GPT-4 extraction and, again, highlights some difficulties with this approach.

This paper aims to contribute to the development of accurate and open systems, at the disposal of the employers as well as the employees, reducing the current knowledge imbalance towards the former, which may have easier access to these systems and the associated knowledge (Adams et al. 2022).

The findings of our study are statistically robust, offering a representative snapshot of the broader UKET database. By critically analysing the performance of GPT-4 in this legal information extraction endeavour, we provide valuable insights into the strengths and limitations of employing LLMs in the legal domain. Furthermore, our research highlights various use cases for extracted information, underscoring its utility in enhancing legal research, improving case management and informing judicial decision-making processes.

## 2 UKET and the English legal system

### 2.1 Fundamental features

In contrast with ordinary courts, which have a wider jurisdiction, the UK for centuries had a number of courts that were local or that dealt with special matters, an example of which are the employment tribunals (Spencer 1989). The UKET is part of the wider judicial system, and one of the three largest tribunals in the greater tribunals system (Courts and Tribunals Judiciary website: https://www.gov.uk/courts-tribunals/employment-tribunal). There are two separate jurisdictions for Employment Tribunals in Great Britain: one for England and Wales, and one for Scotland. Although they each have its own independent judiciary, they share the Employment Tribunals Rules of Procedure 2013, hereinafter referred to as "Rules".

The UKET aims to provide a procedure which is easily accessible, informal, speedy and inexpensive (Harding et al. 2014). One way to achieve this is to require that almost all proceedings are preceded by a conciliation attempt through the Advisory, Conciliation and Arbitration Service (ACAS) (Rules, r. 18). If a conciliation is successful, the employment dispute is not heard by the UKET. Other cases which are not heard by the tribunal are those which are rejected after a preliminary consideration of the claim form and response (Rules, r.

26-28). As a result, UKET decisions do not represent the entirety of the employment dispute outcomes.

Some judgments do not give reasons for the decision. Reasons for judgements may be given orally at the hearing, in which case reasons will not be provided in writing, unless one of the parties so requests within 14 days of the publication of the notice of the judgement (Rules, r. 62(3)). In such scenarios, it is difficult to identify whether the outcome is motivated by procedural or substantive reasons without gathering further information on the specific case.

In contrast with other UK judgments, UKET decisions are clearly structured, not only because there are no dissenting opinions, but also because there is a rule which defines which elements a judgement must contain (Rules, r. 62(5)). Moreover, employment law provisions are often more straightforward than, for example, legal questions decided by the Supreme Court. For this reason, the decisions of the UKET seem to be clearer with a simpler structure compared with the more complex decisions decided in the higher courts.

## 2.2 Relevance of procedure

Even when cases are heard by the UKET, a claim may still fail for procedural reasons. For instance, the UKET may strike out all or part of a claim or response on any of the following grounds: (a) scandalous or vexatious or has no reasonable prospect of success; (b) the manner in which proceedings have been conducted by or on behalf of a party has been scandalous, unreasonable or vexatious; (c) non-compliance with rules or orders of the tribunal; (d) not having been actively pursued; (e) no longer possible to have a fair hearing (Rules, r. 37(1)). Therefore, the fact that a claim has been struck out may be due to substantive issues that go to the heart of the dispute (e.g., having no prospect of success), or to mere procedural issues (e.g., a party not having actively pursued a claim which is meritorious and which would otherwise proceed).

This holds true more generally, as the general outcome of a judgement does not indicate simply who is meritorious but may result from the procedural conduct of the parties. In other words, one cannot deduct from the unsuccessful outcome of a case that the claim had no prospects of success right from the outset. For example, if the claimant does not appear before the tribunal and thereby fails to present the required evidence, the claim can be dismissed (Rules, r. 47). The UKET may also dismiss the claim based on preliminary issues, e.g., if claim is not presented on time, the tribunal is not well constituted or the claimant does not have the required ACAS certificate. Furthermore, when no response has been timely presented, or when any response received has been rejected and no application for a reconsideration is outstanding, or where the respondent has stated that no part of the claim is contested, the tribunal can decide the case on the available material without hearing the respondent (Rules, r. 21). Similarly, if the respondent does not produce evidence (written or orally, by not attending the final hearing), the claims may be found proven if the evidence is unchallenged.

Finally, claims may be withdrawn, for example because the parties reach a private settlement or settle following conciliation by ACAS. Where a claim is withdrawn, the tribunal shall issue a dismissal judgement (Rules, r. 52). Case management orders (e.g., admitting or excluding witness evidence), which are not judgments, may incidentally be part of judgments (Mansfield and Banerjee 2023).

Against this background, procedural issues are relevant because they may be the key reason for a certain outcome. The outcome will depend not only on the facts and the points of law under discussion, but also on the procedural conduct of the parties. Therefore, extracting procedural elements is required for a contextual understanding of the dispute. With this information, one can identify the cases which were decided mainly for procedural reasons, for the purpose of excluding them from the dataset or, contrarily, including them in a useful, contextualised, way.

On the one hand, one may be interested in predicting the likelihood of success of a claim, taking into consideration all procedural developments that may arise. Procedural issues reflect significant obstacles faced by parties in submitting or defending a claim and might be relevant for others who plan to make a procedural application, such as an extension of the time to submit a claim or a response, and, therefore, are interested in considering certain procedural events. Legislators and academics may welcome this approach, as it gives them the possibility of continually monitoring, evaluating and reforming the tribunal system. The payment of tribunal fees is a good example. Currently, the balance between the right to access the tribunals and deterrence of vexatious claims has been established in the following way: claimants do not need pay fees to submit an application, but the claimant may be required to pay a deposit and, exceptionally, the tribunal can award costs against a claimant if it considers that the latter has acted 'vexatiously, abusively, disruptively or otherwise unreasonably' (Deakin et al. 2021). Additional measures could be implemented if, for example, a disproportionate number of vexatious claims are being submitted. The Government could also increase the resources to deal with claims that are burdening the system, or reform the law in areas which display unfairly low levels of success. This approach is also useful for researchers interested in the effective enforcement of employment rights. For such a study, the withdrawal or submission out of time would count as "claimant loses", since, ultimately, the tribunal does not enforce the right.

The Government publishes statistics on the percentages with which the following outcomes have occurred: ACAS conciliated settlements, withdrawn, successful at hearing, unsuccessful at hearing, struck out (not at hearing), default judgement and dismissed at a preliminary hearing. However, these statistics are arranged on a yearly basis, are prone to inaccuracies and do not give details of individual judgments or awards, which prevents more detailed content analysis (Irving 2012).

On the other hand, lawyers may prefer to exclude procedure-based decisions from the dataset, as they may not be interested in integrating procedural elements as inputs for a prediction task. For example, lawyers may not be interested in assessing what happens if they withdraw the claim, as they assume that the procedure is going to be well led by both parties. Instead, they might wish to know the likelihood of succeeding, irrespective of procedural irregularities. From such a perspective, decisions which are motivated by procedural reasons are anomalous and they might be excluded from a prediction dataset. It is then possible to distinguish between procedural predictability (based on a predictive system which includes procedure-based decisions) and substantive predictability (based on a system which excludes procedure-based decisions).

## 3 Information extraction for legal texts

Since the 1950s, legal scholars have been leveraging mathematical and computational tools to analyse a wide range of legal domains. This effort involved using rule-based information extraction methods and statistical extraction approaches. Researchers faced the challenge of either selecting a detailed list of cases and patterns for rule-based analysis or defining a set of features for a statistical system to consider. These features usually included legal and factual criteria, as well as demographic information about the parties involved and the names of the judges (Ashley and Bruninghaus 2006). In relation to employment law, as early as 1965,

Grunbaum and Newhouse (1965) analysed 20 judgments of the US Supreme Court to identify the variables which impacted the outcomes. Similarly, Field and Holley (1982) identified factors which influenced the outcomes of performance appraisal judgments. The authors independently selected the variables, identified the scoring codes for each variable selected and classified each case according to a scoring scheme. Brudney et al. (2000) analysed the extent to which extradoctrinal factors such as political party, gender and professional experience influenced outcomes. Moreover, several studies have been concluded to establish correlations between specific demographic groups and the ability to pursue their employment rights in tribunals. In the US, Schuster and Miller (1984) analysed 153 federal court cases, focusing on age discrimination, and Schultz and Petterson (1992) analysed race and sex discrimination. In the UK, Barnard and Ludlow (2016) investigated whether EU-8 migrant workers were able to enforce their rights by bringing claims before the UKET.

Many of these studies occurred before judgments were published online and, therefore, not only entailed costly journeys to courts and registers, but also required the manual extraction and tagging of the specific elements of the decisions (e.g., jurisdiction, location, gender of the complainant). It is only since February 2017 that UKET decisions are published online. Recently, Blackham (2021) and Irving (2012) concluded a quantitative analysis of employment decisions, but despite having access to online judgments, some tasks still required manual labour.

The limitation of this kind of analysis is that it associates judgment outcomes with pre-selected features, and risks that there are relevant variables, which truly account for the outcome, but which are not taken into account by the model (Ashley and Bruninghaus 2006). The UKET website identifies the "jurisdiction code" of each judgment, such as "breach of contract", "public interest disclosure" and "unfair dismissal". This is helpful because the cases are already sorted by topic. However, despite the multiplicity of jurisdiction codes, many of them are interrelated, such that their use may not be instrumental. Moreover, as explained earlier, the reason for the decision may relate to circumstances that do not relate to the substance of the jurisdiction code.

Lately, some machine learning techniques directly correlate facts with outcomes, without any features having been preselected. These correlations may still be inadequate, because they do not employ abstract legal concepts, drawn from statutes, case law or underlying legal principles, which are normally used to explain a predicted decision (Ashley and Bruninghaus 2006). Moreover, the induced correlations may not correspond to patterns of reasoning which are familiar to legal practitioners (Ashley and Bruninghaus 2006).

Deep learning models based on natural language processing handle the complexity and subtleties of legal language, and, therefore, represent a step-forward. Deep learning models based on natural language processing may, in particular, perform two tasks: extraction and prediction, or a combination of both. Based on relevant information through extensive training on large-scale legal corpora (extraction), these legal-specific language models are considered to be accustomed to the vocabulary and writing styles of legal texts and can be further fine-tuned to facilitate downstream legal judgment prediction tasks (prediction). Accordingly, facilitated by the large-scale datasets, recent years have witnessed a surge in the application of deep learning models to legal judgment prediction.

These models may be used to extract relevant information, with which legal professionals, and potential parties to a court case, can be advised. The challenge of extraction from legal documents is that they can be long, display a complex structure and legal terminology and, in addition, datasets with domain-specific documents are often rare (Collarana et al. 2018).

The retrieval of past judgments which can support a dispute at hand involves an automatic identification of similar situations and suitable statutes or precedent and is a challenge still under exploration. Competitions held recently, such as COLIEE, aimed at exploring the application of natural language processing-based methods and providing benchmarks for the legal case retrieval task. To tackle this challenge, some recent works attempt to model the semantic relationships between legal paragraphs using BERT (Shao et al. 2020).

Beyond extracting information, deep learning models can be used for prediction. As an example, the DataJust project, led by the French Ministry of Justice, aims at offering to the public indicative benchmarks for compensation for personal injuries, with the aim of encouraging out-of-court settlements. The project processes court decisions to extract and exploit data concerning the amounts requested and offered by the parties to the proceedings, the assessments proposed within the framework of procedures for the amicable settlement of disputes and the amounts allocated to victims by the courts (Decree n.° 2020/356 of March 27, 2020 for the creation of an automated processing of personal data called 'DataJust'). This could be extended to employment tribunals, for example, as part of the mandatory conciliation phase. Zhong et al. (2018) introduced TopJudge to address legal judgment prediction using multi-task learning that combines three elements, namely, law articles, charges and terms of penalty. Another notable contribution is the work of Ma et al. (2021) where an end-to-end framework was built to predict legal dispute outcomes with multi-task supervision and multi-stage representation learning. Inspired by the success of pre-trained language models, a number of law-specific pretrained language models have been proposed, such as LegalBERT (Chalkidis et al. 2020) and Lawyer-LLaMA (Huang et al. 2023) for English and Lawformer (Xiao et al. 2021) for Chinese texts. A group of American academics has developed a machine learning application that predicts the outcome of a case at the Supreme Court of the United States (SCOTUS) with an accuracy of 70.2% (Katz et al. 2017). The most extensively discussed application predicts decisions of the European Court of Human Rights (ECHR). This tool uses natural language processing and machine learning to predict whether or not the Court will rule that a particular provision of the ECHR has been violated (Aletras et al. 2016).

Within the framework of the European Union, Recital 40 of the Proposal for an AI Act states that AI systems "intended to assist judicial authorities in researching and interpreting facts and the law and in applying the law to a concrete set of facts" should be qualified as high-risk, not including AI systems "intended for purely ancillary administrative activities that do not affect the actual administration of justice in individual cases". The latter refer to mere back-office tasks, such as the anonymisation of court documents, the communication between personnel and the allocation of resources. As a consequence, AI systems, which allow for researching and interpreting facts and the law, and which might be used by judicial authorities, must be cautiously designed and developed to ensure that models are trained with sufficient accurate and domain specific input data. This applies to both extraction and prediction tools.

If predictive systems are developed and provided by private companies, it is likely that employers, given the superior resources at their disposal, will be in a better position than the employees to access these systems and, therefore, be in a better position to access and understand quantitative information on the chance of tribunal success and the likely level of financial remedies (Adams et al. 2022). Therefore, it would be important to counterbalance this trend with a development of a predictive system to be at the disposal of the general public.

Extraction and prediction still pose some challenges. This paper demonstrates that to effectively develop a useful and accurate prediction model, it is crucial to first analyse the results produced by an extraction model. This paper evaluates the performance of GPT-4 for the extraction of annotations, which proves to be very useful and reliable for few-shot annotations in these settings. GPT-4 can accurately identify the facts, claims and outcomes, as well as the essential reasons for decisions through an extraction task, and thereby challenge the apparent similarity of cases by pinpointing the critical criteria, or features, that dictate the outcome of judgments. With its results, it is possible to identify judgments which should not be part of a prediction model dataset. This paper also highlights the pitfalls that an extraction and prediction model may have, even in a simple scenario such as the UKET, and which must be taken into consideration in the development of any model.

## 4 Information extraction using large language models

In this section, we introduce the data that we conduct our analysis on and the design of the information extraction experiments using a large language model.

### 4.1 Data preparation

#### 4.1.1 The Cambridge Law Corpus

In this research, the experiments and analyses are conducted based on the UKET subset (denoted as $UKET_{ori}$) of the Cambridge Law Corpus (Östling et al. 2023).

The Cambridge Law Corpus is a large-scale dataset for legal AI research, with over 350,000 court cases spanning over 53 courts in the UK. The cases in this dataset cover legal disputes from the 16$^{th}$ century until June 2023, with the majority of cases decided in the late 20$^{th}$ century and the 21$^{st}$ century. The $UKET_{ori}$ subset of the CLC contains 52,339 cases in total, spanning over a time period from 2000 to June 2023. Cases heard at the UKET cover a wide range of issues, such as unfair dismissal, discrimination and breach of contract. Each case file in the $UKET_{ori}$ dataset contains textual transcripts of a court decision, supplemented by a list of metadata such as date of filing, date of decision, place of hearing, jurisdiction codes, judges, claimants, respondents and appearances at the hearing.

#### 4.1.2 Stratified sampling of cases

Out of the 52,339 $UKET_{ori}$ cases, we selected a subset of 260 cases using stratified sampling to make sure that our findings are statistically robust. The selected subset is denoted as $UKET_{sub}$. In the stratified sampling process, all $UKET_{ori}$ cases were first divided into strata using *page count* as a proxy, and then randomly sampled within each stratum. The resulting numbers of sampled $UKET_{sub}$ cases are summarised in Table 1.

| Page count | Number of cases that have been manually checked |
|---|---|
| 1 | 163 |
| 2 | 43 |
| 3 | 9 |
| 4 | 6 |
| 5 | 4 |
| 6 | 3 |
| 7 | 2 |
| 8 | 2 |
| 9 | 2 |
| 10 | 2 |
| 11 | 2 |
| 12 | 2 |
| 13 | 2 |
| 14 | 1 |
| 15 | 1 |
| 16 | 1 |
| 17 | 1 |
| 18 | 1 |
| 19 | 1 |
| 20 | 1 |
| >20 | 11 |

Table 1: The distribution of the 260 cases sampled in this research, categorised using page count.

## 4.2 Information extraction

In this research, we explored the feasibility of utilising a recent large language model for automatic extraction of legal information from case transcripts.

### 4.2.1 Large language models

Large language models (LLMs), such as GPT-4 and its predecessors, represent a significant leap forward in the field of artificial intelligence, particularly in natural language processing (NLP). These models, trained on vast corpora of text from diverse sources, have demonstrated remarkable capabilities in generating coherent, contextually relevant text based on given prompts. Their application spans various domains, from creative writing and chatbots to more specialised fields like medicine and law.

In this research, we focus on a recently released large language model – GPT-4 (OpenAI 2023) – the fourth iteration of the Generative Pre-trained Transformer series developed by OpenAI. The architecture of GPT-4, while adhering to the transformer-based model introduced by Vaswani et al. (2017), incorporates significant enhancements in terms of scale, training data diversity and fine-tuning methodologies. These improvements enable GPT-4 to exhibit a remarkable understanding of context, nuance and the subtleties of human language, surpassing previous models in both qualitative and quantitative evaluations. One of the defining features of GPT-4 is its unprecedented scale, boasting an even larger number of parameters than its predecessor, GPT-3, which itself was a monumental model at its time of release. This increase in scale has been instrumental in improving the model's performance across a variety of NLP benchmarks, making it adept at tasks ranging from text completion and translation to more complex challenges like problem-solving and creative content generation.

### 4.2.2 LLM-aided case annotation

The CLC provides raw texts of the transcripts of UKET decisions. These transcripts usually contain inter-entangled statements about facts provided by parties and their lawyers, reasoning towards a decision, legal statutes and precedents applied to justify the reasoning and final decisions regarding the case outcome and further remedies. In this research, we explored using a large language model, GPT-4 (OpenAI 2023) to automatically disentangle the information in the court transcripts and extract clean statements about eight aspects of legal information, including facts of the case, claims made, references to legal statutes, references to precedents, general case outcome, general case outcome in one of given labels, detailed order and remedies and essential reasons for the decision. We utilised the 32k version of GPT-4 for all experiments in this paper.

An input message to GPT-4 has the following format:
{'role': 'system'; 'content': [PROMPT_TEXT],
'role': 'user'; 'content': [INPUT_TEXT]}

The prompt text refers to the instruction that we provide for GPT-4 to conduct the extraction task. We will discuss in detail our prompt engineering attempts in section 4.3. The input text refers to the raw text of a UKET decision. Examples of the full extraction workflow are illustrated in Annex A.

**4.3 Prompt engineering**

Prompt engineering refers to the process of designing and refining the instructions (or "prompts") given to AI models, particularly in the context of LLMs like GPT-4, to elicit the desired outputs. This process involves crafting questions, statements or scenarios in a way that guides the AI to understand the task at hand and generate accurate, relevant and coherent responses.

Prompt engineering is crucial for harnessing the full potential of LLMs. Well-crafted prompts can significantly improve the quality and relevance of the model's outputs due to the following factors:
1. LLMs are generalist models capable of performing a wide variety of tasks. Prompt engineering allows users to specify the task they want the model to perform, whether it is answering a question, generating text, summarising content, translating languages or even more complex tasks like coding or composing music. The prompt acts as a task descriptor, guiding the model on what is expected.
2. Through prompt engineering, users can provide context or background information that helps the model understand the task better and generate more appropriate responses. This is especially important for tasks that require domain-specific knowledge or nuanced understanding.
3. Prompt engineering is also used to influence the style, tone and format of the model's outputs.

We followed the official guidelines on prompt engineering provided by OpenAI (https://platform.openai.com/docs/guides/prompt-engineering/strategy-test-changes-systematically). In particular, we adopted four of the six proposed strategies, namely: the formulation of clear instructions, the provision of reference text, the splitting of complex tasks into simpler tasks and the systematic testing of changes. As regards the formulation of clear instructions, OpenAI suggests the following approaches: i) include details in the query to get more relevant answers; ii) ask the model to adopt a persona; iii) use delimiters to clearly indicate distinct parts of the input; iv) specify the steps required to complete a task; and v) provide examples.

We applied an iterative development process to find the optimal prompt for our particular purpose. To start with, we established what exactly the model must achieve. What should GPT-4 extract from a judgment? First of all, to get an overall idea of the dispute, it is essential to identify the main *facts*, the *claims* and the *general case outcome*. Employment law issues are governed by contractual provisions, collective bargaining agreements, statutes and case law. Whereas contractual provisions and collective bargaining agreements are context-dependent, anyone interested in understanding a legal question, and in predicting the outcome of the related dispute, must identify the relevant sources, i.e., the *references to legal statutes* and *references to precedents* (Dadgostari et al. 2021). Finally, as argued above, it is important to identify the main *reasons* of the decision, but also to pinpoint whether the *reasons* for deciding a case are procedural or substantive.

Following the second recommendation of OpenAI, we started the prompt by proposing a persona to GPT-4: "You are a legal assistant. Your task is to read through the court decisions that I will send you, and extract the following information for each input". After some iterative testing, we listed the required inputs, using numbers as delimiters to clearly indicate distinct parts of the input, following the third recommendation of OpenAI: 1. *facts* of the case; 2. *claims* made; 3. *references to legal statutes* and regulations; 4. *references to precedents* and other court decisions; 5. *general case outcome*.

Concerning the identification of *facts* and *claims*, it is worth noting that after an initial judgment deciding on the claimant's application, the parties may each submit additional claims or counterclaims, which are then decided in subsequent judgments (e.g., an application for costs). Moreover, often there is a subsequent judgment ascertaining the specific remedies. We observed that GPT-4 was often overinclusive in such subsequent cases, in the sense that it included the description of the original claims as well as the facts connected to those claims, in the *facts* and *claims* output sections, instead of referring exclusively to the facts and claims made in the specific subsequent judgment. This occurred because UKET judgments often report on earlier claims made, without clarifying in a way discernible for GPT-4 that the claim made in the specific proceeding is narrower. The overinclusion of claims had problematic consequences as regards determining the *general case outcome*, given that over-including claims (i.e., including claims which are discussed in the original dispute, but are not under consideration in the decision at hand), may lead to a "partly wins" output when in fact the correct would be "wins" or "loses". We avoided this limitation by including details in the prompt to get more relevant answers (following the first recommendation by OpenAI). After these amendments, the relevant part of the prompt read as follows: "1. *facts* of the case of the specific court decision; 2. *claims* made in the specific court decision and considered in the specific court decision. Do not include any claim which has already been decided in any previous decision."

Asking only for *references to legal statutes* and regulations proved to be too narrow. For example, when the case file referred to the Equality Act 2010, GPT-4's output was that there are "no specific references to legal statutes and regulations". The Equality Act was only captured by GPT-4 if the prompt included the term "rules". Also, GPT-4 did not always identify the procedural tribunal rules. Although it seemed to identify "sections" interchangeably with "articles" or "rules", a more specific and detailed prompt delivered more comprehensive and accurate outputs. Against this background, the final prompt reads as follows: "any references to legal statutes, acts, regulations, provisions and rules, including the specific number(s), section(s) and article(s) of each of them, and including procedural tribunal rules."

Requesting a *general case outcome* sometimes led to unclear outputs. The simple prompt of a *general case outcome* allowed GPT-4 to convey the outputs at its own discretion, which sometimes led to mixing the claims, defences and reasons for the decision in the same output section. Therefore, we specified the steps required to comply with the task, following the fourth recommendation of OpenAI, thus subdividing the prompt into: *general case outcome*; *general case outcome (summarised in one of the four labels)*: "claimant wins", "claimant loses",

"claimant partly wins" and "other"; and *detailed order and remedies*. This *detailed order and remedies* section of the prompt directs GPT-4 to go beyond a mere indication of the *general case outcome* in terms of which party is successful towards the exact content of the decision or order made by the Tribunal. To avoid a result where the reasons for the decision are dispersed over the outcome sections, we added a specific output section for the *reasons*. A more segmented prompt consistently leads to complete and accurate outputs, and the introduction of sections allows for a clearer distinction between facts, claims, outcomes and reasons for the decision.

Regarding the four labels, "claimant wins", "claimant loses", "claimant partly wins" and "other", we noted a particular difficulty with the latter. GPT-4 often employed the label "other" for disputes that were dismissed for procedural reasons. It also produced inconsistent results. For example, it categorised a preliminary judgement, in which the tribunal allowed a case to proceed even though the claim was submitted out of time as "other", whereas similar cases were instead classified as "won" or "lost". To avoid such inconsistencies, we added an explanation to the prompt: "Note that the label 'other' is to be reserved for situations in which the result cannot be determined or where the outcome cannot be described in terms of winning or losing (e.g., an evidence collection)." By providing an example in brackets we applied the fifth recommendation by OpenAI.

When there are multiple parties to the proceedings, the output would be more accurate if GPT-4 managed to specify the outcome for each of them. To achieve this, we introduced a further stipulation: "if there are multiple claimants or respondents, extract the case outcome for each and all of the claimants or respondents separately". Although this specification generally produces the desired detail, it fails in some cases. For example, in a case with 14 claimants, GPT-4 extracted the outcomes for the first two claimants and then stated: "the information for the other claimants can be similarly organised." After testing different prompt variations without solving this problem, we accepted this limitation.

To conclude, the final prompt used in this project reads as follows: "You are a legal assistant. Your task is to read through the court decisions that I will send you, and extract the following information for each input: 1. *facts* of the case of the specific court decision; 2. *claims* made in the specific court decision and considered in the specific court decision. Do not include any claim which has already been decided in any previous decision; 3. any *references to legal statutes*, acts, regulations, provisions and rules, including the specific number(s), section(s) and article(s) of each of them, and including procedural tribunal rules; 4. *references to precedents* and other court decisions; 5. *general case outcome*; 6. *general case outcome summarised* using one of the following four labels - "claimant wins", "claimant loses", "claimant partly wins" and "other". Note that the label "other" is to be reserved for situations in which the result cannot be determined or where the outcome cannot be described in terms of winning or losing (e.g., an evidence collection); 7. *detailed order and remedies*; 8. essential *reasons* for the decision (procedural and substantive). If there are multiple claimants or respondents, extract the case outcome for each and all of the claimants or respondents separately."

**5 Quality check and assessment**

**5.1 Design of quality check**

To evaluate the quality of the GPT-4 extractions, we asked a legal expert to manually check the accuracy of the extracted information and the suitability for each case to be used for a downstream prediction application. The manual check process was supervised by a senior legal expert who helped to resolve questions, revised quality check guidance during the process and made additional random checks. The legal expert is qualified at post-graduate level and the senior legal expert is a law professor. Both have previous experience in annotating legal data in a machine-learning context.

The quality check is divided into two parts. The first part focuses on examining the accuracy of each piece of extracted information by GPT-4. We investigated the eight individual factors that were included in the prompt. The second part concentrates on the suitability of the extracted information for a downstream prediction task. The aim of this second check is to investigate whether the legal information extracted by GPT-4 has the potential to be used for a prediction training dataset.

### 5.1.1. First part of the quality check

For the first part of the quality check, an extraction output is considered accurate if the results of GPT-4 are generally correct against the background of the prompt and the information in the case file provided. In essence, this quality check established whether GPT-4 delivers information that is accurate and useful for a user such as a legal expert or an individual interested in aspects of UKET decisions. We assigned a score of 0 or 1 to the output of each of the eight extraction sections, namely: *facts, claims, references to legal statutes, references to precedents, general case outcome, general case outcome (summarised as one of the four labels), detailed order and remedies* and *reasons*. A score of 0 was assigned if the output was inaccurate, whereas a score of 1 was assigned if it was accurate. The next paragraphs will describe the accuracy test for each of the elements.

- Facts

A judgment comprises different types of facts: substantive facts (i.e., events before the tribunal procedure), procedural facts (i.e., the actions taken in the procedure) and accessory facts (i.e., facts which are neither substantive nor procedural, and which are not under dispute). Our prompt does not provide any instructions as to which facts should be extracted, as previous attempts to induce GPT-4 to distinguish between substantive and procedural facts were unsuccessful. Hence, facts that are extracted by GPT-4 comprise all aspects related to a case, including substantive and procedural facts. We accepted both perspectives as valid in our checks.

If substantive or procedural aspects are accurately described in the *facts* section, the extracted facts are annotated as accurate. The *facts* section of GPT-4's extracted output is annotated as inaccurate when the extracted facts contain an error or when they are incomplete. The latter occurs when the output facts do not reflect the workplace events discussed by the judge to substantiate the applicable legal norms. If the dispute turned on the substantive facts and they are not mentioned in the *facts* section, then the extraction output is classified as inaccurate (see Example 2, Annex B). Where the decision does not summarise the facts at all, GPT-4 was not required to deliver any facts, and an output stating that there are no relevant facts to report was classified as accurate.

For judgments based on procedural reasons rather than substantive reasons (a "procedural decision"), it is considered as acceptable for GPT-4 to omit these procedural events in the *facts* section. This evaluation aligns with our principle that the *facts* section should be limited to substantive facts. Also, procedural decisions often do not touch upon the substantive arguments for or against a claim, as the claim is dismissed due to procedural aspects. In such cases, as mentioned above, we consider the extraction to be accurate if the *facts* section expressly indicates that the case file does not provide information about the facts of the case.

The *facts* section may state "accessory" facts, such as the identity of the parties or the venue in which the proceedings were concluded. If correct, such outputs are classified as accurate. For an illustrative example, see Example 1 in Annex B.

- Claims

If the case file contains information about the claims raised, the claims are expected to be retrieved by GPT-4. The *claims* section extracted by GPT-4 is classified as correct if GPT-4 accurately identifies the claim. When a case file does not contain information about claims made, output stating "the document does not provide details of the specific claims made" is considered accurate. We considered it essential that the absence of claims in the case file is clearly indicated in the output. If a claim is explicitly described by the judge, then any output that mentions a claim without detailing it or asserts that the claim details are not specified in the file, is considered incorrect.

- General case outcome and detailed order and remedies

The *general case outcome* and *detailed order and remedies* sections were classified as correct, if they were accurate and complete. If there were multiple claims, the output was required to give the outcomes for each claim. As judgments vary, there are many different possible combinations of *general case outcome* and *detailed order and remedies*. Just to give some examples illustrating how these two outputs may complement each other: Concerning a declaration claim, the *general case outcome* can be the description of the declaration, while the *detailed order and remedies* section would mention that there are no specific orders or remedies. Concerning claims for damages, the *general case outcome* may mention that the claimant succeeded in relation to specific claims, while the *detailed order and remedies* section could specify that the respondent is ordered to pay a certain sum to the claimant.

- General case outcome (summarised as one of the four labels)

Here, GPT-4 was expected to classify the general case outcome with one of the labels "claimant wins", "claimant loses", "claimant partially wins" and "other". The prompt specified that the label "other" was reserved for situations in which the result cannot be determined or where the outcome cannot be described in terms of winning or losing (e.g., an evidence collection).

- References to legal statutes and precedents

GPT-4's outputs were classified as accurate if they included all statutory and other legal rules and all case-law referred to in the case file. Any imprecision led to a categorisation as inaccurate.

- Reasons

The *reasons* section was classified as correct if GPT-4's extraction accurately identified the facts which determined the outcome. If the judgment was based on the applicable substantive rules (a "substantive decision"), the legal arguments which determined the outcome also had to be included in the *reasons* section. If, on the other hand, the judgment was based on procedural reasons (a "procedural decision"), those procedural aspects had to be identified in the *reasons* section for the output to be considered as accurate.

**5.1.2. Second part of the quality check**

The second part of the quality check focussed on establishing whether the extracted information can be used to construct a computational prediction task. This is to showcase a potential use case for the LLM-extracted information in the computational field. The prediction task aims to predict the likely *outcome* of a UKET case based on the *facts* and the *claims* made in the case. Three sections were examined: the *facts*, the *claims* and the *general case outcome* (labelled as "claimant wins", "claimant loses", "claimant partially wins" or "other") of the information extracted by GPT-4. The relevant quality check is divided into two steps.

First, for a prediction model to effectively predict the most likely decision of a case, it needs clear information about the *facts* and the *claims* as input. If any of these critical sections is missing, the model cannot generate accurate predictions. The *general case outcome (as one of the four labels)* is also necessary for establishing an outcome prediction task, as the outcome labels will be used as gold-standard references to guide model training and testing.

When any of the three sections (i.e., *facts*, *claims* and *general case outcome*) is missing in the GPT-4 extracted information for a case, we consider the case not being suitable to be used for an outcome prediction task. If GPT-4 outputs information about the three sections, but that information does not contain informative content, we also consider the case as not useful for the task. For instance, when GPT-4 outputs, in the *facts* section, that "the document does not provide specific facts of the case", then the case is not useful for the prediction task. Similarly, when the *claims* section mentions that the case file does not provide the claim, this is not suitable for a prediction model. While these outputs were classified as correct (i.e., annotated as "1") in the first part of the quality check, which assessed accuracy, they are now classified with a "0" at this step of the quality check, which assesses usefulness for a prediction model.

Second, the concepts of procedural predictability (which refers to a predictive system which includes procedural actions of the parties) and substantive predictability (which refers to a predictive system which excludes such procedural actions), mentioned above in section 2.1, become relevant here. These case characteristics are distinguished here, since some users may be interested in predicting the likelihood of success of a claim taking into consideration procedural developments that may arise, i.e., they may be interested in the procedural predictability. Others, in contrast, may wish to exclude procedure-based decisions from the dataset, in order to have a system which focusses on substantive prediction. The development of a model of interest to those who prefer the second perspective requires a second step in the quality check. This second step identifies those case files whose *facts* sections are dominated by procedural events.

Therefore, at the second step of this quality check, we evaluated whether the *facts* of a case contain only or are dominated by procedural events and classified those judgments as procedural (annotated as "1"). Such cases allow only for procedural prediction and would need to be excluded from a dataset of a model focusing on substantive prediction. Overall, these are judgments whose extraction outputs are dominated by procedural events in litigation as opposed to events at the workplace.

As explained above, when there is a procedural decision, we did not require GPT-4 to include procedural facts in the *facts* section. In such a case, the *reasons* section will clarify that the judgment was determined by a procedural fact. Such a case will be classified as procedural. If the facts section provides procedural information and is dominated by it, the judgment is also classified as procedural. Examples for such judgments are those in which there is a withdrawal of claim or non-compliance with an order, and for which it is not possible to establish how the tribunal applied the law to the workplace dispute. To the contrary, decisions are annotated as substantive (i.e., with a "0"), if the *facts* mainly concerned workplace events (though they can include procedural matters to a lesser extent). Such judgments can be used for a dataset to prepare a substantive prediction model.

In summary, the second part of the quality check consisted of two steps and resulted in the attribution of two numbers to each judgment, one number for each step of the second part of the quality check. The first step classifies with a "0" those judgments whose *facts, claims* or *general case outcome (as one of the four labels)* sections are either missing or non-informative, because they correctly, but irrelevantly for a prediction task, state that the judgment does not provide information about the facts or claims or because these sections consist only of accessory information. The second step classifies those judgments which are dominated by procedural events with a "1" to facilitate their exclusion from a dataset of a substantive prediction model.

As a result, after the second part of the quality check, each judgment is classified by two annotations. If the first annotation is "0", the judgment does not allow for any kind of prediction. If the first annotation is "1" and the second is also "1", then the judgment is dominated by procedural events, and may be included in a dataset if the users are interested in procedural predictability. The judgments whose first annotation is "1" and whose second annotation is "0", generally allow for prediction and since they are not dominated by procedural events, they allow the system to learn how the UKET applies employment law statutes and regulations as well as case law to workplace conflicts.

## 5.2 Quantitative assessment

The accuracy scores of all extracted aspects for the 260 cases are summarised in Table 2. For each accuracy score, we also computed its corresponding confidence interval with a confidence level of 95%. It can be seen from the table that GPT-4 obtained perfect accuracy (i.e., 100%) for the extraction of *references to legal statutes* and the extraction of *references to precedents*. It also achieved near-perfect accuracy for the extraction of *claims* and *general case outcomes*, with an accuracy score of 0.981 and 0.996 respectively. For the seemingly more challenging tasks of extracting *detailed case outcomes* and constructing *reasons* from legal judgement texts, GPT-4 surprisingly obtained a high accuracy of 0.996 for both tasks. Despite the relatively lower accuracy compared to that of other aspects, the accuracy of GPT-4's extraction of *facts* and *general case outcomes (summarised in one of the four labels)* are still higher than 0.9, highlighting the overall high accuracy of GPT-4's extraction of legal information from diverse legal texts.

Out of the 260 cases, around 47.7% of the cases (i.e., 124 cases) are marked as being suitable to be used for a computational prediction task. Among these cases, 85 cases contain more than 1 page. This suggests that short court judgements usually do not contain sufficient information about case facts, claims or outcomes for the judgement texts to be effectively used for a downstream prediction task. We also analysed the accuracy scores and corresponding confidence intervals for the 124 cases. The results are listed in the third column in Table 2. Similar to those of the 260 cases, all extracted aspects yielded an accuracy higher than 0.9. The accuracy for *facts* for the 124 cases has dropped compared to that of all 260 cases, whilst the accuracy for *general case outcomes (summarised in one of the four labels)* has increased.

| Aspect of extraction | Accuracy and CIs (all 260 cases) | Accuracy and CIs (124 cases suitable for prediction) |
|---|---|---|
| (1) facts | 0.942 ± 0.028 | 0.919 ± 0.033 |
| (2) claims | 0.981 ± 0.017 | 0.976 ± 0.019 |
| (3) references to legal statutes | 1.000 | 1.000 |
| (4) references to precedents | 1.000 | 1.000 |
| (5) general outcomes | 0.996 ± 0.008 | 0.992 ± 0.011 |
| (6) general outcomes in one of four labels | 0.912 ± 0.034 | 0.952 ± 0.026 |
| (7) detailed outcomes | 0.996 ± 0.008 | 0.992 ± 0.011 |
| (8) reasons | 0.996 ± 0.008 | 0.992 ± 0.011 |

Table 2. Accuracy and confidence intervals for the eight factors extracted by GPT-4. We report the evaluation results on all the 260 cases (the second column) and on the 124 cases (the third column) that are suitable for a downstream prediction task.

## 5.3 Qualitative assessment of information extraction

Generally, the extraction outputs were very accurate. Only the *general case outcome* given as one of four labels had a higher degree of inaccuracy than the other sections. Annex A contains two examples of the input case contents and GPT-4's extracted outputs, the first one extracted from a short case file, and the second one extracted from a longer case file. These examples demonstrate the accuracy of GPT-4 for extraction purposes. The evaluation of GPT-4's outputs in terms of accuracy requires contextual consideration, not only of the operational mechanisms of the system, but also of the information contained in the case files. This is particularly important for the evaluation of the *outcome* sections.

To start with, when claims are for the payment of certain amounts, UKET judgments do not necessarily refer to the amounts of the claim as originally submitted. Consistently, when identifying the outcome of a certain claim, GPT-4 does not consider the original amount of the claim or the subitems which are comprised within the same claim (e.g., holidays or wages). As a consequence, GPT-4 classifies the outcome as "claimant wins", and not as "claimant partially wins", in a situation in which a claimant is only partly successful (e.g., it was not possible to show evidence of some elements of the damage and, therefore, the total amount of the initial claim was not granted). We classified this as accurate, since the UKET case file did not give the amount claimed. Independent of this, GPT-4 was expected to use the "partially wins" label when the claimant submits multiple claims, and some are lost and the others won.

Since GPT-4 adopts the view of the claimant, when there are counterclaims and both the claimants and the respondents' claims are upheld, GPT-4 classifies the case as "partially wins". We accepted this approach as correct, since it applied a defensible logic and was applied consistently throughout. When respondents submit a counterclaim, they are still referred to as "respondents" in the case text, although, substantively, they are the claimants as regards the counterclaim. Since GPT-4 always considers the claimants to be the original claimants, if the respondents succeed on their own counterclaims, GPT-4 labels the outcome as "claimant loses". Applying the above principles contextually, we classified this as accurate.

Some cases are complex and raise annotation doubts even for human experts. For example, when a claimant wins subject to a reduction for contributory fault, one could argue, on the one hand, that because the amounts petitioned are not considered, but just the overall success of the claim, the outcome should be labelled as "claimant wins". On the other hand, considering that the contributory fault is connected to a counterclaim of the respondent, which is also upheld, the label "claimant partially wins" might be better suited. Indeed, GPT-4 took inconsistent approaches to this issue, sometimes using the label "claimant wins" and sometimes the label "claimant partially wins". Whereas any of the views could be classified as accurate, we classified inconsistence as a mistake and only accepted one of these approaches as correct (the "claimant partially wins" label).

GPT-4 displayed a surprisingly good grasp of the outcome label "other". In a correction case, GPT-4 correctly refused to apply a label, which indicates avoidance of "hallucination" or improper inference where the data did not support a clear outcome. Similarly, in a case which was settled out of court and the claims were stayed to allow for the implementation of the settlement terms, GPT-4 correctly output the label "other". As regards the label "other", labelling mistakes of GPT-4 were not a result of wrong classifications per se, but rather of applying an inconsistent logic across all outputs for similar cases. For example, a difficult labelling situation arises when the Tribunal does not decide on the claim and, instead, allows it to proceed to a final hearing. Should the outcome be classified as "claimant wins", because the claim is not struck out, as the respondent requested, or as "other", because the claim is not finally decided? Again, we required GPT-4 to apply one approach consistently across all outputs and treated the "other" label as correct. Finally, there was a labelling inconsistency of GPT-4 across cases where the claims were withdrawn. Although the majority of the withdrawal cases were accurately labelled as "claimant loses", in the sense that the claimant did not succeed, there were 16 cases in which GPT-4 labelled those cases as "other", which we treated as a mistake. The reason for GPT-4's inconsistency might be that a withdrawal does not correspond to a decision that directly disadvantages the claimant. Again, this point represents a labelling inconsistency across the dataset, rather than an inaccuracy per se.

**5.4 Discussion on the usage for a prediction task and procedural facts**

In the first part of the quality check, we evaluated whether GPT-4's extractions were accurate. Having an accurate information extraction system is valuable *per se,* and it offers a strong foundation for developing a wide range of downstream applications. One such downstream application may be, for example, a case outcome prediction model that predicts future workplace dispute outcomes based on facts and claims made. Given the strong impact that this type of downstream application is likely to have on society, it is important to build strong safeguards around its development and to be cautious about its limitations. For that reason, in the second step of the quality check, we evaluated whether the information extracted by GPT-4 (in particular, *facts, claims* and *labelled outcomes*) can be used to establish a computational downstream task that predicts future workplace dispute outcomes based on facts and claims.

In the second part of the quality check, the human annotators were active actors, because GPT-4 did not previously select the judgments which allow for substantive prediction. Instead, the human annotators analysed and selected which of the judgments would allow for substantive or procedural prediction, based on the outcomes produced by GPT-4 for the extraction task. The second part of the quality check is a task of content analysis. We looked, first, at the *facts* and *claims* sections and established whether they are informative or irrelevant. Secondly, we looked at the *facts* section and determined whether the information contained therein is mainly procedural or substantive.

Our manual evaluation revealed that approximately half of the cases are viable for incorporation into a predictive analysis task, with a noticeable trend showing that most suitable cases extend beyond a single page. This pattern suggests that the information extracted by GPT-4 holds significant potential for constructing predictive tasks and for the training and evaluation of predictive models. The observation that longer cases typically offer more substantive content underscores their value as both input and output in computational tasks. This finding is particularly relevant for those interested in using LLM-curated information for downstream computational analysis. It also highlights the importance of case length as an indicator of informational richness, suggesting that more extended documents may provide a deeper, more nuanced basis for analysis.

There is a relative limitation to the second part of the quality check and, overall, to the substantive predictability of judgments. There are judgments which result from the appreciation of the applicable substantive rules and the application of those rules to the workplace dispute. Those are substantive judgments, which allow for substantive predictability, in contrast to procedural judgments. The relevant second part of the quality check identifies the judgments which are mainly based on procedural events, with the aim of allowing to exclude those cases from a prediction task. However, as mentioned in Section 2.1, some outcomes, even if based on a substantive appreciation of the dispute, may be strongly influenced by the fact that the respondent does not produce evidence (written or orally; by not attending the final hearing), or by the fact that the respondent did not present a response (Rules, r. 21). These judgments are still based on a substantive appreciation of the law and of the facts and are, indeed, substantive judgments, but are strongly influenced by procedural events. Those cases would not be pinpointed by the second task of the second part of the quality check, although they allow for a limited substantive predictability. Take, for example, the application of r. 21. The lack of response or contest by the respondent, being a procedural issue, is not necessarily indicated in the GPT-4 output *facts* section. Out of all the cases of our manually verified dataset, 26 cases applied r. 21. Of those, 9 cases mentioned r. 21 in the *facts*, the *references to legal statutes* and *reasons* sections, 10 cases mentioned r. 21 just in the *references to legal statutes* and 7 cases mentioned it in the *references to legal statutes* and in the *reasons* section, but not in the *facts* section. In other words, the majority of the cases only include r. 21 in the references. There is no lack of extraction accuracy in any of these scenarios. This limitation inherently affects all predictive systems across various legal sectors, given that the parties must present the evidence to the judges, and that certain legal rights need to be invoked by the interested parties. Hence the concept of "procedural truth", so familiar to law students (in opposition to the actual truth, sometimes neglected). Nevertheless, this is a relevant consideration to take into account, especially considering that the AI Act classifies these systems as high-risk systems.

Another limitation of which users of these systems should be aware is that a prediction model whose training dataset includes only tribunal case files does not take into consideration all the elements that judges dispose of, such as the information included in the claim and response forms. This seems an insurmountable obstacle where the dataset does not contain the relevant information. Also, taking into account the judgments of a specific tribunal or court leaves behind many disputes which were settled amicably out-of-court. In addition, two or more claimants may submit their claims using the same claim form if their claims give rise to common or related issues of fact or law (Rules, r. 9). Furthermore, in a situation in which there are two different claims filed, involving the consideration of the same facts and legal principles, the UKET may consider both cases at the same time, which is referred to as the claims being "conjoined", or "heard together", and is regulated as a "group litigation order" in r. 19.22 of the Civil Procedure Rules. These discussions outline the inherent complication of establishing a predictive task based on the Employment Tribunal data constructed in this research, which should be carefully thought through by designers and users of these systems.

## 6 Limitations

Our study provides valuable insights into the practice of employing LLMs for extracting information from UK Employment Tribunal case documents, highlighting the efficiency and potential accuracy of this approach. The adoption of GPT-4 for this extraction task is beneficial in terms of time and budget efficiency. The extraction results are arguably satisfactory according to the manual quality check conducted by legal experts. Nonetheless, our findings come with an acknowledgment of certain areas for improvement.

**Distinguishing procedural and substantive facts in the prompt**. GPT-4 does not always mention procedural elements in the *facts* section. This leads to incomplete factual descriptions. In those cases, the *reasons* section may often (but not always, as the discussion of r. 21 above demonstrates) add procedural details, for example by clarifying that there was a withdrawal, or that the respondent did not appear or present a response, which might have determined or influenced the outcome. However, the outputs of the *reasons* section cannot be included in the training dataset of a prediction model, since it would lead to data leakage. Our prompt specifies that the *reasons* section must include procedural and substantive elements, specification which is not included as regards the *facts* section. Perhaps a similar clarification for the *facts* section could result in more complete factual descriptions.

**Access to the actual facts and claims of the cases**. This limitation is only relevant for the discussions on the prediction task that we showcased in section 5.4. The facts and claims produced by GPT-4 are extracted from tribunal decision transcripts, i.e., the judges' written decisions at the end of the proceedings. Since the judges know the result of the case at this stage of the process, the texts they write may inherently contain biased information. For example, sentiment words in the judges' statements might implicitly reveal their inclinations towards certain decisions. When trained on such data, the models might incorporate such factors when making predictions related to case outcomes. While we believe that the LLM-extracted facts and claims have the potential to be used as a practical substitute to the actual facts and claims submitted by parties before a hearing, we acknowledge that this approach could potentially introduce information biases at the input stage of the prediction task. In subsequent research, we will explore alternative methods of identifying facts and claims to better approximate the original submissions to the court, thus fostering a more realistic modelling of judgment prediction.

## 7 Conclusion

The intersection of LLMs and legal AI is a burgeoning field, with significant potential for transforming traditional legal workflows, enhancing access to justice and providing insights that were previously unattainable due to the prohibitive costs of manual legal analysis. In our experiments, GPT-4 has achieved strong results in extracting key information from decisions of UKET. The first task required GPT-4 to extract information on facts, claims, general and detailed outcomes, legal rules and case law applied as well as the relevant reasons. The second task prepared a prediction experiment aiming to predict case outcomes based on facts and claims raised. As this technology continues to evolve, its integration into legal practices is expected to deepen, heralding a new era of legal AI where LLMs play a critical role in shaping the future of the legal profession and research.

**References**

Adams Z, Adams-Prassl A and Adams-Prassl J (2022). Online tribunal judgments and the limits of open justice. Legal Studies 42(1):42-60. https://doi.org/10.1017/lst.2021.30

Aletras N, Tsarapatsanis D, Preoţiuc-Pietro D and Lampos V (2016) Predicting judicial decisions of the European Court of Human Rights: A natural language processing perspective. PeerJ Computer Science, 2:93. https://doi.org/10.7717/peerj-cs.93

Ashley K and Bruninghaus S (2006) Computer Models for Legal Prediction. Jurimetrics 46(3):309-352.

Banerjee L and Mansfield G (2023) Blackstone's Employment Law Practice. Oxford University Press, Oxford. https://doi-org.ezp.lib.cam.ac.uk/10.1093/oso/9780192887788.001.0001

Barnard C and Ludlow A (2016) Enforcement of Employment Rights by EU-8 Migrant Workers in Employment Tribunals. Industrial Law Journal 45(1):1-28.

Bernsohn D, Semo G, Vazana Y, Hayat G, Hagag B, Niklaus J, Saha R and Truskovskyi K (2024) LegalLens: Leveraging LLMs for Legal Violation Identification in Unstructured Text. arXiv preprint arXiv:2402.04335. https://doi.org/10.48550/arXiv.2402.04335

Blackham A (2021) Enforcing rights in employment tribunals: insights from age discrimination claims in a new 'dataset.' Legal Studies 41(3):390-409. https://doi.org/10.1017/lst.2021.11

Brudney J, Schiavoni S and Merritt DJ (1999) Judicial Hostility Toward Labor Unions? Applying the Social Background Model to a Celebrated Concern. Ohio State Law Journal 60(5):1675-1771. https://ir.lawnet.fordham.edu/faculty_scholarship/166

Chalkidis I, Fergadiotis M, Malakasiotis P, Aletras N and Androutsopoulos I (2020) LEGAL-BERT: The Muppets straight out of Law School. In Findings of the Association for Computational Linguistics: EMNLP 2020, pp 2898-2904. https://doi.org/10.18653/v1/2020.findings-emnlp.261

Collarana D, Heuss T, Lehmann J, Lytra I, Maheshwari G, Nedelchev R, Schmidt T and Tridevi P (2018) A Question Answering System on Regulatory Documents. Legal Knowledge and Information Systems 313:41-50. https://doi.org/10.3233/978-1-61499-935-5-41

Dadgostari F, Guim M, Beling PA, Livermore MA and Rockmore DN (2021) Modelling law search as prediction. Artificial Intelligence and Law, 29, pp 3-34. https://doi.org/10.1007/s10506-020-09261-5

Deakin S, Adams Z, Barnard C and Butlin S (2021) Deakin and Morris' Labour Law. Hart Publishing, Oxford.

Field HS and Holley WH (1982) The Relationship of Performance Appraisal System Characteristics to Verdicts in Selected Employment Discrimination Cases. The Academy of Management Journal 25(2):392-406. https://doi.org/10.5465/255999

Grunbaum WF and Newhouse A (1965) Quantitative analysis of judicial decisions: some problems in prediction. Houston Law Review 3(2):201-220.

Harding C, Ghezelayagh S, Busby A and Coleman N (2014) Findings from the Survey of Employment Tribunal Applications 2013. Research Series No. 177, Department for Business Innovation and Skills of the United Kingdom Government. https://assets.publishing.service.gov.uk/media/5a749243e5274a410efd0aa6/bis-14-708-survey-of-employment-tribunal-applications-2013.pdf. Accessed 28 February 2024.

Huang Q, Tao M, An Z, Zhang C, Jiang C, Chen Z, Wu Z and Feng Y (2023) Lawyer LLaMA Technical Report. arXiv preprint arXiv:2305.15062. https://doi.org/10.48550/arXiv.2305.15062

Irving LC (2012) Challenging Ageism in Employment: An Analysis of the Implementation of Age Discrimination Legislation in England and Wales. Dissertation, University of Coventry.

Jayakumar T, Farooqui F and Farooqui L (2023) Large Language Models are legal but they are not: Making the case for a powerful LegalLLM. In Proceedings of the Natural Legal Language Processing Workshop 2023, pp 223-229. https://doi.org/10.48550/arXiv.2311.08890

Katz DM, Bommarito MJ II and Blackman J (2017) A general approach for predicting the behavior of the Supreme Court of the United States. PLoS ONE 12(4): e0174698. https://doi.org/10.1371/journal.pone.0174698

Ma L, Zhang Y, Wang T, Liu X, Ye W, Sun C and Zhang S (2021) Legal judgment prediction with multi-stage case representation learning in the real court setting. In Proceedings of the 44th International ACM SIGIR Conference on Research and Development in Information Retrieval, pp. 993-1002. https://doi.org/10.1145/3404835.3462945

OpenAI (2023) GPT-4 technical report. arXiv preprint arXiv:2303.08774. https://doi.org/10.48550/arXiv.2303.08774

Östling A, Sargeant H, Xie H, Bull L, Terenin A, Jonsson L, Magnusson M and Steffek F (2024) The Cambridge Law Corpus: A Corpus for Legal AI Research. Advances in Neural Information Processing Systems, 36. https://doi.org/10.48550/arXiv.2309.12269

Parizi AH, Liu Y, Nokku P, Gholamian S and Emerson D (2023) A Comparative Study of Prompting Strategies for Legal Text Classification. In Proceedings of the Natural Legal Language Processing Workshop 2023, pp 258-265. https://doi.org/10.18653/v1/2023.nllp-1.25

Schultz V and S Petterson S (1992). Race, Gender, Work, and Choice: An Empirical Study of the Lack of Interest Defense in Title VII Cases Challenging Job Segregation. University of Chicago Law Review 59(3):1073-1181. https://ssrn.com/abstract=1025534

Schuster M and Miller C (1984) An Empirical Assessment of the Age Discrimination in Employment Act. ILR Review 38(1):64.74. https://doi.org/10.1177/001979398403800107

Shao Y, Mao J, Liu Y, Ma W, Satoh K, Zhang M and Ma S (2020) BERT-PLI: Modelling Paragraph-Level Interactions for Legal Case Retrieval. Proceedings of the Twenty-Ninth International Joint Conference on Artificial Intelligence (IJCAI-20): 3501-2507. https://doi.org/10.24963/ijcai.2020/484

Spencer JR (1989) Jackson's Machinery of Justice. Cambridge University Press, Cambridge. https://doi.org/10.1017/CBO9780511560071

Vaswani A, Shazeer N, Parmar N, Uszkoreit J, Jones L, Gomez AN, Kaiser Ł and Polosukhin I (2017) Attention is all you need. Advances in Neural Information Processing Systems, 30. https://doi.org/10.48550/arXiv.1706.03762

Xiao C, Hu X, Liu Z, Tu C and Sun M (2021) Lawformer: A Pre-trained Language Model for Chinese Legal Long Documents. arXiv preprint arXiv:2105.03887. https://doi.org/10.48550/arXiv.2105.03887

Zhong H, Guo Z, Tu C, Xiao C, Liu Z and Sun M (2018) Legal judgment prediction via topological learning. In Proceedings of the 2018 conference on Empirical Methods in Natural Language Processing, pp 3540-3549. https://doi.org/10.18653/v1/D18-1390

**Annex A**

This annex provides two examples illustrating the overall workflow and results of extracting legal information from UKET cases using GPT-4.

**Example 1. Case 3328920/2017**

*Prompt to GPT-4:*
You are a legal assistant. Your task is to read through the court decisions that I will send you, and extract the following information for each input: 1. facts of the case of the specific court decision; 2. claims made in the specific court decision and considered in the specific court decision. Do not include any claim which has already been decided in any previous decision; 3. any references to legal statutes, acts, regulations, provisions and rules, including the specific number(s), section(s) and article(s) of each of them, and including procedural tribunal rules; 4. references to precedents and other court decisions; 5. general case outcome; 6. general case outcome summarised using one of the following four labels – 'claimant wins', 'claimant loses', 'claimant partly wins' and 'other'. Note that the label 'other' is to be reserved for situations in which the result cannot be determined or where the outcome cannot be described in terms of winning or losing (e.g., an evidence collection); 7. detailed order and remedies; 8. essential reasons for the decision (procedural and substantive). If there are multiple claimants or respondents, extract the case outcome for each and all of the claimants or respondents separately.

*Input text to GPT-4:*
See text in
https://assets.publishing.service.gov.uk/media/5a7d7e5e40f0b64fe6c240d5/Mr_Y_Mfunda_v_Swaay_Child_and_Adolescent_Services_Ltd_3328920_17__Full.pdf.

*GPT-4's response:*
1. **Facts of the case**: The case involves Mr Y Mfunda (Claimant) and Swaay Child and Adolescent Services Limited (Respondent). The Respondent failed to present a response to the claim. The Claimant was owed wages and claimed for unauthorised deduction from wages and breach of contract related to mileage expenses.
2. **Claims made in the specific court decision**: The Claimant claimed for unauthorised deduction from wages and breach of contract related to mileage expenses.
3. **References to legal statutes, acts, regulations, provisions and rules**: The case refers to Rule 21 of Schedule 1 to the Employment Tribunals (Constitution and Rules of Procedure) Regulations 2013.
4. **References to precedents and other court decisions**: There are no references to precedents or other court decisions in the provided text.
5. **General case outcome**: The Respondent was ordered to pay the Claimant £6,690.75 (gross) in compensation for the unauthorised deduction from wages. The claim for breach of contract related to mileage expenses was dismissed as the Claimant is still employed and does not have the right to bring a claim for breach of contract.
6. **General case outcome summarised**: Claimant partly wins.
7. **Detailed order and remedies**: The Respondent is ordered to pay the Claimant £6,690.75 (gross) in compensation for the unauthorised deduction from wages. The hearing listed for 14 May 2018 is cancelled.
8. **Essential reasons for the decision**: The Respondent failed to present a response to the claim. The Claimant was found to be owed wages, hence the award for unauthorised deduction from wages. The claim for breach of contract was dismissed as the Claimant is still employed and does not have the right to bring a claim for breach of contract.

**Example 2. Case 2301070/2018**

*Prompt to GPT-4:*
The same prompt as used in Example 1.

*Input text to GPT-4:*
See text in
https://assets.publishing.service.gov.uk/media/5ce5071ced915d2482756121/Mr_A_Martin_v__1_Southwark_Council_2_The_Governing_Body_of_Evangelina_Hospital_School_-_2301070_2018.pdf

*GPT-4's response:*
**1. Facts of the case**: The claimant, Mr A Martin, is a teacher employed by Southwark Council and the Governing Body of Evangelina Hospital School. He alleges that he was required to work more than the statutory limit of 1265 directed hours per academic year without additional pay, amounting to unauthorised deductions from his wages. He also alleges that he made protected disclosures about this issue, which led to detrimental treatment.
**2. Claims made in the case**: The claimant made two main claims. Firstly, he claimed for unauthorised deductions from wages due to being required to work beyond the statutory limit of directed hours without additional pay. Secondly, he claimed that he suffered detrimental treatment as a result of making protected disclosures about this issue.
**3. References to legal statutes, acts, regulations, provisions and rules**: The case refers to the Employment Rights Act 1996, specifically section 13 relating to unauthorised deductions from wages. It also refers to the School Teachers' Pay and Conditions Act 1991 and the Employment Tribunals (Constitution and Rules of Procedure) Regulations 2013, particularly rules 2, 37 and 39.
**4. References to precedents and other court decisions**: The case of Agarwal v Cardiff University is mentioned, which confirmed that a Tribunal may determine the construction of a contract of employment in an unauthorised deduction from wages claim. The case of Chesterton Global v Nurmohamed is also referred to in relation to whether the disclosures made were in the public interest.
**5. General case outcome**: The Tribunal struck out the claimant's claim for unauthorised deductions from wages, finding that it had no reasonable prospect of success. The claimant's protected disclosure claim was partly struck out, with one alleged disclosure having no reasonable prospect of success. The Tribunal refused the respondents' applications for a strike out or deposit order on other grounds. The claimant's application to add two additional respondents was also refused.
**6. General case outcome summarised using one of the following four labels**: 'Claimant partly wins'.
**7. Detailed order and remedies**: The Tribunal did not order any remedies as it struck out the claimant's claim for unauthorised deductions from wages and partly struck out his protected disclosure claim. The Tribunal also refused the respondents' applications for a strike out or deposit order on other grounds and the claimant's application to add two additional respondents.
**8. Essential reasons for the decision**: The Tribunal found that the claimant had no legal entitlement to pay for directed hours worked over the statutory limit of 1265 per year, and therefore his claim for unauthorised deductions from wages had no reasonable prospect of success. The Tribunal also found that one of the claimant's alleged protected disclosures had no reasonable prospect of success as it was a disclosure to a third party that no one from the respondents saw. The Tribunal refused the respondents' applications for a strike out or deposit order on other grounds and the claimant's application to add two additional respondents, finding that the balance of prejudice tipped in favour of the respondents.

**Annex B**

This Annex contains extractions from GPT-4's output for particular case files identified by case number. The original case file can be found on the official website of the UKET, https://www.gov.uk/employment-tribunal-decisions. For each case, GPT-4 has produced outputs for each of the eight sections. Due to limitations of space, we merely present the sections which are relevant to illustrate the points mentioned in the main text.

Example 1: case number 3305262/2021. Illustration of a judgment whose facts section containing accessory facts was classified as accurate.
**1. Facts of the case:** The case involves Ms Tanika Harrison (claimant) and Network Rail Infrastructure Limited (respondent). The case was heard remotely via telephone due to COVID-19 restrictions.
**2. Claims made in the specific court decision:** The claimant, Ms Tanika Harrison, had made a claim for unlawful deduction of pay. However, this claim was dismissed upon withdrawal.

Example 2: case number 2602938/2022. Illustration of a substantive judgment whose facts section was classified as inaccurate because it did not refer to the substantive elements of the dispute. Specifically, the facts section lacks the essential aspect, the unauthorised deductions.
**1. Facts of the case:** The claimant, Mr G Steadman, issued a claim in the Midlands East Employment Tribunals on 10 December 2022 against the respondent, Riatex Limited. The respondent failed to present a valid response on time.
**2. Claims made:** The claimant alleged that the respondent made unauthorised deductions from his wages.
**8. Essential reasons for the decision:** The decision was made in accordance with Rule 21 of the Employment Tribunals Rules of Procedure 2013 due to the respondent's failure to present a valid response on time. The respondent was found to have made unauthorised deductions from the claimant's wages.

Example 3: case number 4107496/2014. Illustration of a judgment whose claims section was classified as incorrect. When the claims are not provided in the text, this must result clearly from the output, or otherwise the section unduly hints that there is an identified claim when there was not one.
**2. Claims made in the specific court decision:** The claimant, Mr W Mollan, had made a claim against the respondents, Arrow XI Limited, but later withdrew it.

Example 4: case number 1302495/2017. This case demonstrates that when the reasons for decision are missing, it might still be possible to make inferences from the facts, claims and outcomes sections. In the quality check concerning the prediction task, this case was classified as suitable for a prediction dataset and not dominated by procedural aspects.
**1. Facts of the case:** Miss M Gugiu, the claimant, was an employee of Vv S Diner Limited, the respondent. The respondent did not attend the hearing. The claimant alleged that she suffered an unlawful deduction from wages, was not paid holiday pay, did not receive notice pay, and did not receive a statement of terms and conditions.
**2. Claims made in the specific court decision:** The claimant made four claims against the respondent:
- Unlawful deduction from wages (£1627.50)
- Unpaid holiday pay (£196.35)
- Unpaid notice pay (£300.00)
- Not receiving a statement of terms and conditions (£600.00)
**5. General case outcome:** The court ruled in favor of the claimant, Miss M Gugiu. The respondent, Vv S Diner Limited, was ordered to pay a total of £2723.85 to the claimant.

**6. General case outcome summarised using one of the following four labels:** "Claimant wins".

**8. Essential reasons for the decision (procedural and substantive):** The court decision does not provide specific reasons for the decision. However, it can be inferred that the court found the respondent to be in violation of employment laws regarding wage deductions, holiday pay, notice pay, and provision of terms and conditions.